\documentclass[sigconf]{acmart}
\usepackage{amsmath}
\usepackage{algorithm}
\usepackage{algorithmic}
\usepackage{graphicx}
\usepackage{booktabs}
\usepackage{subcaption}
\usepackage{multirow}
\usepackage{xcolor}
 
\AtBeginDocument{%
  }

\setcopyright{acmlicensed}
\copyrightyear{2026}
\acmYear{2026}
\setcopyright{cc}
\setcctype{by}
\acmConference[KDD '26]{Proceedings of the 32nd ACM SIGKDD Conference on Knowledge Discovery and Data Mining V.2}{August 09--13, 2026}{Jeju Island, Republic of Korea}
\acmBooktitle{Proceedings of the 32nd ACM SIGKDD Conference on Knowledge Discovery and Data Mining V.2 (KDD '26), August 09--13, 2026, Jeju Island, Republic of Korea}
\acmDOI{10.1145/3770855.3818066}
\acmISBN{979-8-4007-2259-2/2026/08}

\begin{document}

\title{Beyond Linear Dynamics: Neural Bilinear Dynamical Models for Time Series Forecasting}


\author{Mengzhou Gao}
\affiliation{%
  \institution{Hangzhou Dianzi University}
    \city{Hangzhou}
  \country{China}
  }
\email{mzgao@hdu.edu.cn}

\author{Huangqian Yu}
\affiliation{%
  \institution{ Hangzhou Dianzi University}
    \city{Hangzhou}
  \country{China}
  }
\email{232270034@hdu.edu.cn}

\author{Pengfei Jiao}
\affiliation{%
  \institution{Hangzhou Dianzi University}
    \city{Hangzhou}
  \country{China}
  }
\email{pjiao@hdu.edu.cn}
\authornote{Corresponding author.}

\renewcommand{\shortauthors}{Mengzhou Gao, Huangqian Yu, and Pengfei Jiao}

\begin{abstract}
  Time series in real-world applications are often generated by nonlinear dynamical systems, making accurate forecasting challenging. Existing approaches that explicitly model system dynamics typically rely on linear assumptions or Koopman-based linearizations, which may inadequately capture complex nonlinear behaviors and lead to error accumulation in long-horizon prediction.  To address this limitation,  we propose the \textbf{N}eural \textbf{B}ilinear \textbf{D}ynamical \textbf{M}odel (NBDM), which models nonlinear system dynamics through a bilinear  latent dynamical formulation. Specifically, NBDM leverages Koopman theory to lift the original nonlinear dynamics into a higher-dimensional latent space, where a bilinear dynamical model is constructed to characterize state evolution. To mitigate the approximation error introduced by bilinear representations, we further incorporate a parameterized error compensation term. Within this formulation, control inputs are explicitly integrated into the dynamics, using auxiliary variables when available and learned feedback signals otherwise.  To handle scenarios with missing control inputs, we design a memory-enhanced controller that infers latent controls through multiplicative interactions between historical states and control signals. Experiments on five real-world datasets demonstrate that NBDM consistently outperforms competitive baselines in both given-control and missing-control settings, particularly for multi-step and long-horizon forecasting.
\end{abstract}

\ccsdesc[500]{Computing methodologies~Neural networks}
\ccsdesc[500]{Applied computing~Forecasting}

\keywords{Koopman, Neural ODE, Bilinear, Time series}


\maketitle

\section{Introduction}

Time series forecasting is a fundamental task with widespread applications in critical domains such as finance~\cite{berger2026deep,cho2025diffolio}, healthcare~\cite{srikummoon2026time,moon2025graph}, 
transportation~\cite{kong2024spatio,lan2022dstagnn}, 
climate science~\cite{yi2018deep,blanchard2025record} and energy management~\cite{li2025multiscale,zhou2025hierarchical}. In the real world, observed time series often originate from underlying nonlinear dynamical systems whose governing equations are seldom fully known, making their accurate modeling a significant challenge.

To capture these complex nonlinear dependencies, deep learning approaches such as Recurrent Neural Networks (RNNs)~\cite{weng2023decomposition,fan2025pdg2seq,weng2025pattern}, Convolutional Neural Networks (CNNs)~\cite{wu2020connecting,tang2023spatio,chen2024signed}, and more recently Transformer-based~\cite{nietime,liu2023itransformer,jiang2023pdformer} architectures have become prominent. These models excel at learning rich temporal and structural representations directly from data and have demonstrated powerful predictive capabilities. However, they often function as implicit black-box predictors, lacking an explicit dynamical system structure. Consequently, their predictions can be data-hungry, less interpretable, and sometimes physically inconsistent, particularly when extrapolating over multiple prediction steps. In contrast, traditional data-driven forecasting methods have frequently relied on linear dynamical models such as autoregressive models and linear state-space models, due to their analytical tractability and computational efficiency. Nonetheless, these linear approximations often fail to capture essential nonlinearities, leading to systematic biases and degraded performance, especially  in multi-step  forecasting  settings.

A promising middle ground has emerged in recent years through Koopman theory~\cite{koopman1931hamiltonian}. This framework provides a principled way to  represent nonlinear dynamics through linear evolution  in a lifted function space. By mapping the original state into  a higher-dimensional latent representation where the dynamics evolve approximately linearly, Koopman-based methods enable the application of linear systems theory to nonlinear problems.  In practice, neural networks are used to learn finite-dimensional lifted representations, resulting in architectures known as Koopman autoencoders or deep Koopman models.  However, enforcing strictly linear evolution in the latent space can be restrictive when modeling complex nonlinear and coupled interactions, limiting the expressive power of Koopman-based models in real-world systems. A natural extension is to introduce bilinear dynamics, which preserve much of the analytical structure of linear systems while explicitly modeling multiplicative interactions between states and control variables.

Furthermore, real-world dynamical systems are often influenced by exogenous control inputs or external drivers that interact multiplicatively with the system state. For example, traffic flow evolution depends on both the current traffic state and external or auxiliary factors, whose interactions jointly determine future system behavior. Most existing deep forecasting models either ignore such inputs or incorporate them only additively, thereby failing to capture the essential state–control interactions inherent in many physical and engineered systems.

To address these challenges, we propose the Neural Bilinear Dynamical Model (NBDM), a novel deep learning framework that explicitly models nonlinear system dynamics through a bilinear formulation in the Koopman space. Bilinear systems represent a natural middle ground between linear and fully nonlinear dynamics. They are linear in state and control separately but capture their multiplicative interactions, enabling a richer class of dynamical behaviors while retaining considerable analytical structure. Our approach leverages Koopman theory to identify a lifting transformation that maps original observations into a latent space where the dynamics evolve bilinearly. Crucially, we introduce a parameterized error term to compensate for residual modeling errors arising from both the lifted representation and the bilinear structural approximation. This allows the model to adaptively correct residual errors that cannot be captured by the bilinear form. Furthermore, our bilinear formulation explicitly introduces multiplicative terms between the latent state and control variables, enabling a principled and expressive way to model these interactive effects. 

To handle practical settings where external  control inputs are not always available  across different datasets,  we further design a memory-enhanced controller that infers latent control signals from historical dynamics. Specifically, in addition to the classical  system state-based linear term, the controller incorporates multiplicative interaction terms between historical states and  previously observed or inferred control signals. This design enables a unified modeling framework that can naturally accommodate both settings with explicit control inputs and those where such inputs are unavailable.

We evaluate NBDM on five real-world datasets spanning climate, air quality, and traffic forecasting domains.  Experimental results demonstrate that NBDM consistently outperforms existing baseline methods under both given-control and missing-control settings. Notably, NBDM shows significant advantages in extended-term prediction tasks over models based on strict linear system assumptions, highlighting the importance of capturing  nonlinear dynamical effects through bilinear structure.

In summary, the main contributions of this work are as follows:
\begin{itemize}
\item We propose the Neural Bilinear Dynamical Model (NBDM), a novel deep learning framework based on bilinear dynamics that effectively captures complex nonlinear dependencies.
\item  We introduce a unified control input modeling strategy that supports both given-control and missing-control settings by leveraging available external variables and inferring latent control signals via a memory mechanism.
\item We conduct extensive experiments on five real-world datasets, showing that NBDM consistently outperforms state-of-the-art baselines in forecasting accuracy under varied control input conditions, especially for extended-term predictions.
\end{itemize}

\section{Related Work}

\subsection{Time Series Forecasting}

Time series forecasting has been extensively studied, leading to a wide range of deep learning approaches for modeling complex spatial-temporal dynamics.

Early approaches predominantly utilized RNNs. DCRNN~\cite{li2018dcrnn_traffic} incorporates bidirectional diffusion convolution into a gated recurrent unit (GRU), modeling time series as a graph diffusion process. DDGCRN~\cite{weng2023decomposition} extends this by introducing spatial-temporal embeddings and dynamic signals to explicitly encode time-varying node relationships. PDG2Seq~\cite{fan2025pdg2seq} further incorporates a period-aware module and a period dynamic graph convolutional GRU to mine fine-grained spatial-temporal patterns. Convolution-based architectures subsequently gained popularity by replacing the
recurrent computation with parallel temporal convolutions, enabling more efficient training and inference. 
STGCN~\cite{yu2018spatio} combines one-dimensional causal convolutions with graph convolutions to preserve temporal causality while enabling efficient parallelization. 
Graph WaveNet~\cite{wu2019graph} expands the receptive field exponentially using dilated causal kernels and learns an adaptive adjacency matrix. MTGNN~\cite{wu2020connecting} further integrates multiple kernel sizes per channel and reweights graph edges based on node similarities.  Continuous-time dynamical models have also been explored for spatial-temporal forecasting.  MTGODE~\cite{jin2022multivariate} models the joint evolution of hidden states across space and time through a coupled ordinary differential equation (ODE). SGODE~\cite{chen2024signed} further augments this ODE framework with signed Laplacians to suppress spurious correlations between opposing time series directions. More recently, Transformer-based architectures have become dominant due to their strong capacity for long-range dependency modeling. PDFormer~\cite{jiang2023pdformer} introduces a spatial–temporal dual-path block with delay-aware shifts to capture congestion propagation lags. iTransformer~\cite{liu2023itransformer} treats variates as tokens to explicitly model multivariate correlations. SimpleTM~\cite{chen2025simpletm} employs  wavelet tokenization and  generalized self-attention to learn multi-scale temporal and inter-channel dependencies, offering a lightweight yet competitive baseline. TimeMixer~\cite{wang2023timemixer} effectively integrates multi-scale information across both historical analysis and future prediction through a decomposable multi-predictor mixing mechanism.

Although these methods achieve strong empirical performance, most existing forecasting approaches rely heavily on highly nonlinear black-box architectures and lack explicit modeling of the underlying system dynamics.

\subsection{Learning Dynamics with Koopman Operator}
Koopman theory~\cite{koopman1931hamiltonian} provides a promising framework for modeling nonlinear temporal dynamics through linear evolution in a lifted latent space. 

DeepKoopman~\cite{lusch2018deep} and DeepDMD~\cite{yeung2019learning}  employ deep neural networks to learn observable functions spanning Koopman-invariant subspaces and approximate the Koopman operator using linear transformations. Consistent dynamic Koopman AE~\cite{azencot2020forecasting} further design an autoencoder framework for forecasting fluid dynamics.
Subsequent studies extend Koopman learning to more complex temporal systems. K-Forecast~\cite{lange2021fourier} leverages Koopman theory to handle nonlinearity in temporal signals and proposes optimizing data-dependent bases for long-term time series forecasting. ~\citet{li2021deep} constructs a loss function by integrating multi-scale basis functions to learn the Koopman-invariant subspace and reconstruction operator of nonlinear multi-scale dynamical systems from coarse scale data. ~\citet{fan2022learning} introduces a data-driven Koopman embedding approach, which lifts the nonlinear system to a linear manifold and directly parameterizes stable linear operators to achieve unconstrained stable learning. KNF~\cite{wang2023koopman} utilizes deep neural networks to learn the coefficients of the linear Koopman space and the selected measurement function, adapting to constantly changing data distributions. Recent works further explore Koopman modeling for non-stationary and graph-structured time series. 
Koopa~\cite{liu2023koopa} hierarchically disentangles time-invariant and time-variant components of non-stationary series and learns their Koopman embeddings and evolution operators through multilayer perceptrons.
KoopGCN~\cite{wang2024koopman} integrates GCN with Koopman theory to address the prediction challenges posed by non-stationary and unprecedented time series patterns.

Despite their effectiveness, existing Koopman-based approaches often rely on simplified linear evolution assumptions, limiting their ability to capture highly complex nonlinear dynamics in real-world multivariate time series.

\section{Background}

\subsection{System Dynamics Models}

\subsubsection{Nonlinear System}
A general nonlinear dynamical system with control inputs is represented in discrete time as:
\begin{equation}
\label{eq:nonlinear}
    x_{t+1} = \mathbf{F}(x_t, u_t),
\end{equation}
where \( \mathbf{F}: \mathbb{R}^n \times \mathbb{R}^m \to \mathbb{R}^n \) denotes a nonlinear flow map, \( x_t \in \mathbb{R}^n \) represents the system state, and \( u_t \in \mathbb{R}^m \) denotes the control input. Such systems can exhibit complex behaviors and are widely used to represent physical, biological, and engineering processes with strong coupling, uncertainty, or saturation effects.


\subsubsection{Linear System}
A linear dynamical system with control inputs is expressed as:
\begin{equation}
    x_{t+1} = \mathbf{A} x_t + \mathbf{B} u_t,
\end{equation}
where \( \mathbf{A} \in \mathbb{R}^{n \times n} \) and \( \mathbf{B} \in \mathbb{R}^{n \times m} \) are system matrices. Linear systems satisfy the superposition principle and provide a tractable modeling framework. They are often used as local approximations of nonlinear dynamics around operating points.


\subsubsection{Bilinear System}
A bilinear system extends linear dynamics by introducing multiplicative interactions between the state and control input:
\begin{equation}
    x_{t+1} = \mathbf{A} x_t + \mathbf{B} u_t + \sum_{i=1}^{m}u_t^i \mathbf{N}_i x_t ,
\end{equation}
where \( \mathbf{N}_i \in \mathbb{R}^{n \times n} \) captures the coupling between the state and the \( i \)-th control component \( u_t^i \). Bilinear systems can model a wide range of phenomena where control inputs modulate system dynamics, providing greater expressiveness than linear systems while remaining more structured than general nonlinear dynamics. This makes them a structured intermediate representation between linear and general nonlinear dynamics.


\subsection{Koopman Operator Theory}
The Koopman operator theory~\cite{koopman1931hamiltonian} provides a linear operator framework for  analyzing  nonlinear dynamical systems by lifting their evolution into an infinite-dimensional observable space. 

Given an observable function $\phi: \mathbb{R}^n \rightarrow \mathbb{R}$, the Koopman operator $\mathcal{K}$ acts on the space of observables and advances them forward in time. Formally, its action is defined as:
\begin{equation}
\mathcal{K}\phi = \phi \circ \mathbf{F},
\end{equation}
where $\circ$ denotes function composition.

Consider the nonlinear dynamical system in Eq.~\eqref{eq:nonlinear}, whose state evolution is governed by the nonlinear flow map $\mathbf{F}(x_t, u_t)$, where $u_t$ is treated as an exogenous input. The Koopman operator is then defined as $\mathcal{K}\phi(x_t, u_t) = \phi \circ \mathbf{F}(x_t, u_t)$. Applying the Koopman operator to this system yields the evolution of observables:
\begin{equation}
\label{eq:4}
\phi(x_{t+1}) = (\mathcal{K}\phi)(x_t, u_t) = \phi \circ \mathbf{F}(x_t, u_t).
\end{equation}

In practice, finite-dimensional approximations of the Koopman operator induce structured latent dynamics, which can recover linear or bilinear forms depending on the underlying system structure.




\subsection{Problem Formulation}
Let $\mathbf{X}\in \mathbb{R}^{N\times L}$ denote a multivariate time series consisting of $N$ variables observed over $L$ time steps. Specifically, we define $x^i\in \mathbb{R}^{L}$ as the $i$-th variable for all time steps, and $x_t\in \mathbb{R}^{N}$ as the $t$-th time step for all series. Given a sequence of $\mathcal{T}$ historical observations $\mathbf{X}_{t+1:t+\mathcal{T}}\in \mathbb{R}^{N\times \mathcal{T}}$, our goal is to learn a mapping function $\mathbf{F}$ to predict future $\mathcal{T}^{'}$ steps,
\begin{equation}
[\mathbf{X}_{t+1},\cdots ,\mathbf{X}_{t+\mathcal{T}}]\xrightarrow{\mathbf{F}}[\mathbf{X}_{t+\mathcal{T}+1},\cdots ,{\mathbf{X}_{t+\mathcal{T}+\mathcal{T}^{'}}]}.
\end{equation}

\begin{figure*}[htbp]
\centering
\includegraphics[width=2\columnwidth]{
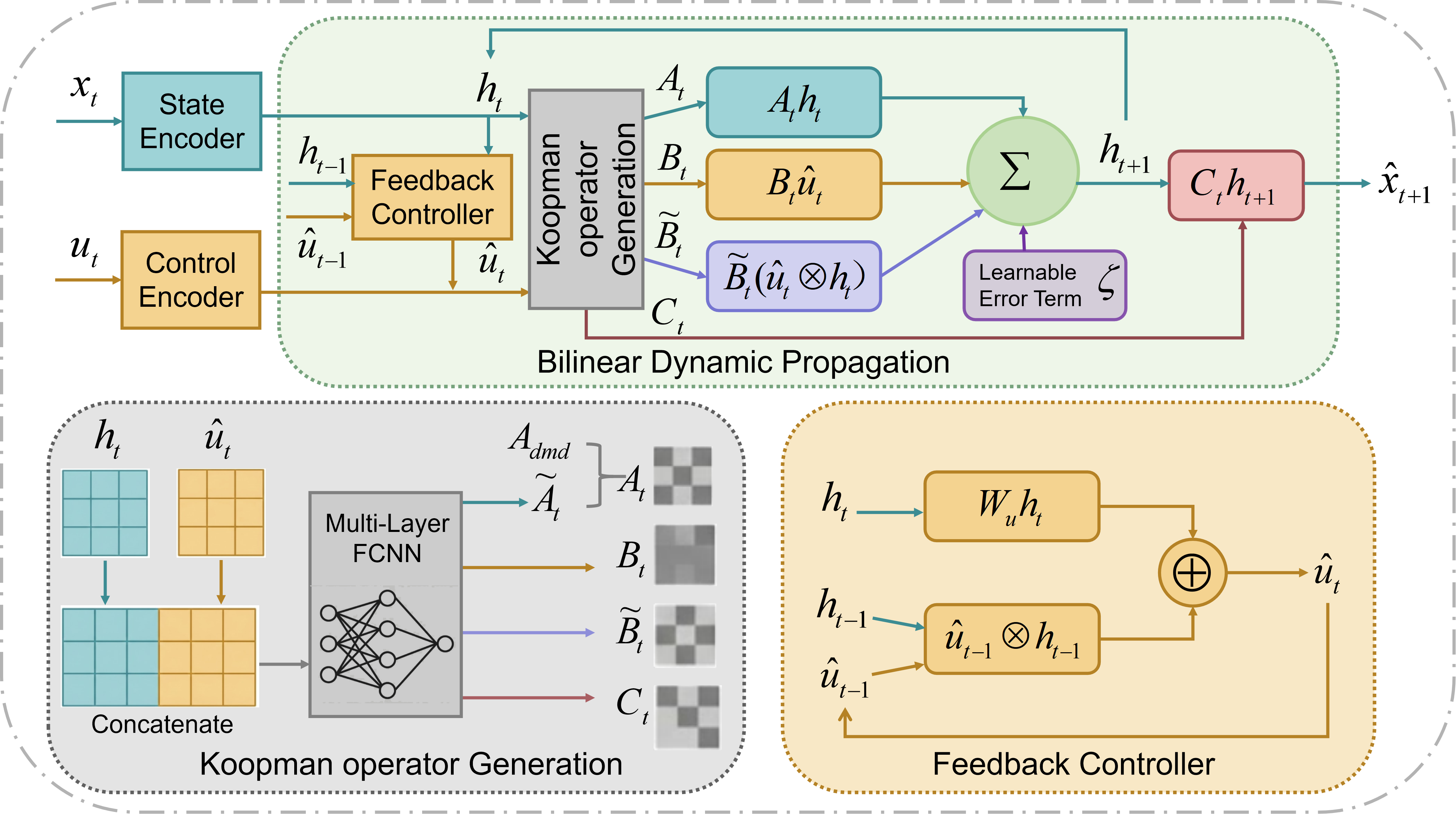}
\caption{The overall framework of NBDM. The system states $x_t$ and available control inputs $u_t$ are separately mapped into the Koopman space via dedicated encoders, yielding their corresponding lifted representations $h_t$ and $\hat{u}_t$. In the bilinear dynamics propagation block, $h_t$ and $\hat{u}_t$ are used to generate Koopman operator matrices and are then  iterated to produce the final multi‑step predictions. When control inputs are unavailable in the dataset, a memory-enhanced  feedback controller is employed to infer implicit control signals  from historical state–control interactions.}
\label{fig:Model}
\end{figure*}

\section{Overall Framework}

In this section, we present a comprehensive description of the proposed Neural Bilinear Dynamical Model (NBDM) and its components. 
The overall framework of our method, shown in Fig.~\ref{fig:Model}, consists of three core components: bilinear dynamic propagation, Koopman operator generation, and feedback controllers. Specifically, system states and control inputs are initially mapped into a latent Koopman space. Subsequently, hidden system states and control inputs are used to generate Koopman operator matrices. For datasets without explicit control inputs, we design a memory-enhanced feedback control strategy  to infer latent control signals. The predicted states are then obtained through iterative bilinear dynamic propagation in a closed-loop manner.

\subsection{Bilinear Dynamic Propagation}

The system states $x_t \in \mathbb{R}^n$  and control inputs $u_t \in \mathbb{R}^m$ are  first embedded into a latent representation  through two separate networks, which are typically two-layer fully connected neural networks. 
\begin{equation}
    h_t = \phi _{\theta_x}(x_t), 
\end{equation}
\begin{equation}
    \hat{u}_t = \phi _{\theta_u}(u_t), 
\end{equation}
where $h_t \in \mathbb{R}^{N\times d_h}$ denotes the latent state representation and $\hat{u}_t \in \mathbb{R}^{N\times d_u}$ denotes the latent control input. Here, $\theta_x$ and $\theta_u$ are learnable  parameters of the corresponding embedding networks. 



The bilinear Koopman model has been shown to be more effective in capturing the coupling effects between control inputs and system states, thereby providing a more expressive representation of underlying nonlinear dynamical systems~\cite{strasser2026overview}. Motivated by this observation and the Koopman operator design in~\cite{strasser2026safedmd}, we construct the following bilinear Koopman-based latent dynamics model:
\begin{align}
h_{t+1} &= \mathbf{A}_t \odot h_t + \mathbf{B}_t \hat{u}_t + \mathbf{\tilde{B}}_t(\hat{u}_t\otimes h_t) + \zeta, \\
\hat{x}_{t+1} &= \mathbf{C}_t h_{t+1}. 
\label{eq:output}
\end{align}
where $\mathbf{A}_t$, $\mathbf{B}_t$, and $\mathbf{\tilde{B}}_t$ are state evolution, control-driven dynamics, and bilinear interaction, respectively. $\odot$ and $\otimes$ denote Hadamard and Kronecker product, respectively. $\zeta$ is a learnable residual term that compensates for approximation errors not captured by the structured dynamics and is further detailed in Sec.~\ref{sec:error_term}. 

The first two terms $\mathbf{A}_t\odot h_t + \mathbf{B}_t\hat{u}_t$ capture the linear dynamics of the system states and control inputs. The third term $\mathbf{\tilde{B}}_t(\hat{u}_t \otimes h_t)$ introduces a bilinear component to explicitly model the multiplicative coupling between the system states and the control inputs. The final term $\zeta$ serves as an approximator to capture any residual complex dynamics not covered by the linear and bilinear parts. 

For multi-step forecasting tasks, we adopt a multi-step rollout strategy, where the model propagates the state from $t$ to $t+1$ and repeats this process in a closed-loop fashion.

\subsection{Koopman Operator Generation}
The core of Koopman theory lies in finding an infinite-dimensional linear operator that can globally linearize a nonlinear system. However, in practical applications, a fundamental challenge is how to construct a finite-dimensional, high-fidelity approximation of this operator from limited data via parameterization. The quality of this approximated operator directly determines the predictive performance of the model.

Initially, we compute an $\mathbf{A}_{dmd} \in \mathbb{R}^{N\times N}$ from the entire training data $\mathbf{X}\in \mathbb{R}^{N\times L}$ using the Dynamic Mode Decomposition (DMD)~\cite{schmid2010dynamic} method , which is a widely used data-driven method for extracting dynamic features from time-series data. Specifically, given system state snapshots $\mathbf{X} = [x_1, x_2, \dots, x_{L-1}]$ and $\mathbf{X}' = [x_2, x_3, \dots, x_{L}]$, DMD seeks a linear operator $\mathbf{A}_{dmd}$ such that 
$x_{t+1} \approx \mathbf{A}_{dmd} x_t$, obtained via 
\begin{equation}
 \mathbf{A}_{dmd} \approx \mathbf{X}' \mathbf{X}^\dagger,  
\end{equation}
where $\dagger$ denotes the pseudo-inverse. This approach is model-free and computationally efficient, as $\mathbf{A}_{dmd}$ remains static throughout the training phase.

Then, the Koopman operator  components $\mathbf{\widetilde{A}}_t \in \mathbb{R}^{N\times d_h}$, $\mathbf{B}_t \in \mathbb{R}^{N\times d_h\times d_u}$, $\widetilde{\mathbf{B}}_t \in \mathbb{R}^{N\times d_h\times d_hd_u}$, and decoder matrix $\mathbf{C}_t \in \mathbb{R}^{N\times d_h}$ are generated from the concatenation of latent system states and latent control inputs:
\begin{equation}
z_t = \mathrm{MLP}([h_t \Vert \hat{u}_t]) ,
\label{eq:param_generation}
\end{equation}
where $z_t$ is reshaped into multiple structured components corresponding to different interaction terms:
\begin{equation}
    [\mathbf{\widetilde{A}}_t, \mathbf{B}_t, \widetilde{\mathbf{B}}_t, \mathbf{C}_t] = \text{split}(z_t, [d_h, d_h d_u, d_h^2 d_u, d_h]).
    \label{eq:param_split}
\end{equation}

Finally, the Koopman operator matrices $\mathbf{A}_t \in \mathbb{R}^{N\times d_h}$ is obtained by interacting $\mathbf{\widetilde{A}}_t$ with a predefined $\mathbf{A}_{dmd}$:

\begin{equation}
    \mathbf{A}_t = \mathbf{A}_{dmd}\mathbf{\widetilde{A}}_t .
    \label{eq:A_computation}
\end{equation}

This parameterization combines a fixed DMD-based global dynamical prior with adaptive state-dependent operator generation, enabling the model to capture both global system structure and local dynamical variations.


\subsection{Feedback Controller Design}
The controller is not merely appended as an auxiliary input feature but instead plays an active role in governing the gain and direction of state evolution at each time step. This makes it a key component for modeling the complex nonlinear dependencies in controlled dynamical systems.


We consider two scenarios depending on whether control inputs are observed or need to be inferred from system dynamics. \textbf{(i) Given control inputs.} 
When control inputs are available in the dataset, they can be directly projected into the latent space through an embedding layer, without additional modeling.  \textbf{(ii) Missing control inputs.} When control inputs are unavailable, we design a method to learn and infer the latent control inputs directly from the system states as well as historical system states and control inputs. The formulation for feedback controller is as follows:
\begin{equation}
    \hat{u}_t =
    \begin{cases}
        \mathbf{W_u} h_t, & \text{if } t = 1, \\
        \mathbf{W_u} h_t + \mathbf{W_{hu}}(h_{t-1} \otimes \hat{u}_{t-1}), & \text{if } t > 1.
    \end{cases}
    \label{eq:control_input_piecewise}
\end{equation}
where $\mathbf{W_u} \in \mathbb{R}^{d_h \times d_u}$ and $\mathbf{W_{hu}} \in \mathbb{R}^{d_h \times d_ud_h}$ are learnable parameters. Compared with a standard linear controllers $\hat{u}_t =\mathbf{W_u} h_t$, the proposed controller introduces a memory-based state–control interactive feedback mechanism. This structure enables the controller not only to respond rapidly to changes in system states but also to dynamically adjust and refine its current output based on the control effectiveness from the previous time step.

\subsection{Learnable Error Term}
\label{sec:error_term}
Within our Koopman operator-based modeling framework, two primary sources of error exist: (i) the approximation error arising from the observables mapping the nonlinear system to a lifted linear space, and (ii) the inherent truncation error due to the finite-dimensional approximation of the infinite-dimensional Koopman operator. 

To explicitly model and compensate for these inevitable systematic errors, we introduce a learnable error term  $\zeta \in \mathbb{R}^{N\times d_h}$ into our formulation. This allows the model to autonomously identify and fit the residual dynamics not fully captured by the linear Koopman evolution.

\subsection{Loss Function}

By iteratively rolling out single-step predictions forward and stacking them across the time horizon, we obtain a multi-step prediction sequence $\mathbf{\hat{X}} \in \mathbb{R}^{N\times \mathcal{T}^{'}}$. Given $\mathbf{X} \in \mathbb{R}^{N\times \mathcal{T}^{'}}$ as the ground-truth  of the future $\mathcal{T}^{'}$ traffic observations, the training objective is defined using the mean squared error (MSE)  loss as follows:
\begin{equation}
\mathcal{L}_{MSE}=\frac{1}{N\mathcal{T }^{'}}\displaystyle\sum_{i=1}^{N}\displaystyle\sum_{t=1}^{\mathcal{T}^{'}}\left| \hat{\mathbf{X}}_t^i - \mathbf{X}_t^i \right|_2^2.
\end{equation}

\section{Experiments}

We evaluate the proposed model on five real-world datasets and compare it with  representative  baseline models under consistent experimental settings. We further conduct a case study  to analyze the learned feedback controller, followed by ablation study and parameter sensitivity analysis. The code is available at \url{https://github.com/mzgaooo/NBDM}.

\subsection{Datasets and Baselines}
We evaluate the proposed model on five real-world datasets covering climate, air quality, and traffic forecasting. The datasets include both settings with and without exogenous control inputs, as summarized in Tab.~\ref{tab:dataset}.

\begin{table}[H]
\centering
\Large
\caption{Summary of datasets used in experiments, including control input availability.}
\resizebox{0.48\textwidth}{!}{
\label{tab:dataset}
\renewcommand{\arraystretch}{1.08}
\begin{tabular}{ccccc}
\toprule
Datasets & Sensors & Time Steps & Data Type & Control Inputs\\
\midrule
Temperature &  18 & 8,784 & Climate & No\\
Seoul PM$_{2.5}$ & 25 & 17,520 & Air Quality & No\\
PeMS-Bay & 325 & 52,116 & Traffic Speed & No\\
PeMS04 & 307 & 16,992 & Traffic Flow & Speed \\
PeMS08 & 170 & 17,956 & Traffic Flow & Speed \\
\bottomrule
\end{tabular}}
\end{table}

Several representative baselines are selected for comprehensive comparison, which are grouped into three categories.

\begin{table*}[t]
\centering
\caption{Performance comparison of baseline methods and ablation study of NBDM on three datasets without available control inputs. Bold and underline denote the best and second-best results, respectively.}
\label{tab:results1}

\renewcommand{\arraystretch}{1.06}
\small
\setlength{\tabcolsep}{3.6pt}

\begin{tabular*}{\textwidth}{@{\extracolsep{\fill}}lll|cccc|cccc|cccc}
\toprule

\multirow{2}{*}{\shortstack[c]{Type}} &
\multirow{2}{*}{\shortstack[c]{Model}} &
\multirow{2}{*}{\shortstack[c]{Metric}} &
\multicolumn{4}{c|}{Seoul PM$_{2.5}$} &
\multicolumn{4}{c|}{Temperature} &
\multicolumn{4}{c}{PeMS-Bay} \\
\cmidrule(lr){4-7}
\cmidrule(lr){8-11}
\cmidrule(lr){12-15}

& & &
1 & 4 & 7 & 10
& 1 & 4 & 7 & 10
& 1 & 4 & 7 & 10\\
\midrule

\multirow{4}{*}{Koopman}
& \multirow[c]{2}{*}{DeepKoopman} & RMSE & 3.83 & 4.52 & 5.62 & 6.92 & 1.79 & 3.71 & 5.76 & 7.99 & 4.54 & 4.81 & 5.17 & 5.54 \\
&              & MAE  & 2.30 & 2.88 & 3.70 & 4.59 & 1.34 & 3.04 & 4.99 & 6.96 & 2.62 & 2.75 & 2.92 & 3.11 \\
& \multirow[c]{2}{*}{Koopa}        & RMSE & 0.45 & 1.21 & 2.23 & 3.45 & 1.54 & 4.06 & 5.76 & 6.45 & 2.71 & 4.16 & 5.24 & 6.15 \\
&              & MAE  & 0.28 & 0.80 & 1.51 & 2.34 & 1.07 & 3.05 & 4.57 & 5.30 & 1.25 & 1.82 & 2.25 & 2.66 \\
\midrule

\multirow{6}{*}{Graph}
& \multirow[c]{2}{*}{DDGCRN}  & RMSE & 0.26 & 1.04 & 2.07 & 3.20 & 1.51 & 3.67 & \underline{5.02} & 5.89 & \underline{1.73} & \underline{3.34} & \underline{4.05} & \underline{4.46} \\
&         & MAE  & 0.20 & 0.76 & 1.51 & 2.33 & 1.03 & 2.97 & 4.32 & 4.88 & \underline{0.89} & \underline{1.62} & 1.95 & 2.34 \\
& \multirow[c]{2}{*}{PDG2Seq} & RMSE & 0.22 & 1.06 & 2.17 & 3.35 & 1.41 & 3.62 & 5.08 & 5.78 & \underline{1.73} & 3.40 & 4.18 & 4.59 \\
&         & MAE  & \underline{0.15} & 0.74 & 1.52 & 2.36 & 1.05 & 3.97 & 4.13 & 4.85 & 0.91 & 1.64 & \underline{1.90} & \underline{2.16} \\
& \multirow[c]{2}{*}{SFGDE}   & RMSE & 1.18 & 0.99 & \underline{1.85} & \underline{2.83} & 2.41 & 4.01 & 6.26 & 6.88 & 1.83 & 3.61 & 4.60 & 5.24 \\
&         & MAE  & 0.38 & 0.71 & 1.33 & \underline{2.05} & 1.56 & 3.12 & 4.86 & 5.35 & 0.95 & 1.70 & 2.16 & 2.50 \\
\midrule

\multirow{6}{*}{Sequence}
& \multirow[c]{2}{*}{iTransformer} & RMSE & \underline{0.21} & \underline{0.91} & 1.91 & 3.07 & 1.37 & \underline{3.52} & 5.19 & \underline{5.75} & 1.85 & 3.84 & 5.10 & 6.06 \\
&              & MAE  & \underline{0.15} & \underline{0.63} & \underline{1.32} & 2.12 & 0.99 & 2.83 & 4.05 & \underline{4.43} & 0.92 & 1.69 & 2.18 & 2.60 \\
& \multirow[c]{2}{*}{SimpleTM}     & RMSE & 0.24 & 1.08 & 2.18 & 3.45 & \underline{1.25} & 3.95 & 5.51 & 5.85 & 1.89 & 3.97 & 5.27 & 6.26 \\
&              & MAE  & 0.17 & 0.75 & 1.51 & 2.38 & 0.98 & \underline{2.80} & 4.42 & 4.72 & 0.94 & 1.72 & 2.22 & 2.64 \\
& \multirow[c]{2}{*}{TimeMixer}   & RMSE & 0.77 & 1.49 & 2.44 & 3.55 & 1.29 & 3.56 & 5.33 & 5.76 & 1.87 & 3.54 & 4.32 & 4.82 \\
&              & MAE  & 0.51 & 0.99 & 1.71 & 2.58 & \underline{0.95} & 2.86 & 4.31 & 4.58 & 0.96 & 1.69 & 2.05 & 2.31 \\
\midrule

\multirow{2}{*}{Ours}
& \multirow[c]{2}{*}{NBDM} & RMSE & \textbf{0.20} & \textbf{0.78} & \textbf{1.73} & \textbf{2.72} & \textbf{1.08} & \textbf{3.44} & \textbf{4.91} & \textbf{5.15} & \textbf{1.67} & \textbf{3.29} & \textbf{3.99} & \textbf{4.37} \\
&      & MAE  & \textbf{0.13} & \textbf{0.50} & \textbf{1.19} & \textbf{1.91} & \textbf{0.89} & \textbf{2.76} & \textbf{4.01} & \textbf{4.16} & \textbf{0.83} & \textbf{1.56} & \textbf{1.85} & \textbf{2.06} \\
\midrule

\multirow{6}{*}{Ablation}
& \multirow[c]{2}{*}{\hspace{0.3cm} -w/o BT} & RMSE & 0.35 & 1.08 & 1.99 & 3.20 & 1.48 & 4.15 & 5.69 & 5.97 & 1.82 & 3.46 & 4.23 & 4.62 \\
&      & MAE  & 0.24 & 0.77 & 1.44 & 2.33 & 1.14 & 3.38 & 4.78 & 5.00 & 0.94 & 1.72 & 2.01 & 2.23 \\
& \multirow[c]{2}{*}{\hspace{0.3cm} -w/o ET} & RMSE & 0.38 & 0.98 & 1.93 & 3.01 & 1.64 & 3.57 & 5.03 & 5.85 & 1.79 & 3.40 & 4.18 & 4.63 \\
&      & MAE  & 0.29 & 0.69 & 1.37 & 2.21 & 1.17 & 2.84 & \underline{4.04} & 4.76 & 0.94 & 1.64 & 1.95 & \underline{2.16} \\
& \multirow[c]{2}{*}{\hspace{0.3cm} -w/o PC} & RMSE & 0.35 & 0.94 & 1.93 & 3.03 & 1.31 & 4.11 & 6.03 & 6.98 & 1.77 & 3.36 & 4.08 & 4.50 \\
&      & MAE  & 0.25 & 0.67 & 1.43 & 2.26 & 0.99 & 3.40 & 4.94 & 5.76 & 0.95 & 1.65 & 1.97 & 2.17 \\
\bottomrule
\end{tabular*}
\end{table*}

\textbf{(i) Koopman-based models.}
DeepKoopman~\cite{lusch2018deep} learns nonlinear encoders that map observations into a latent space where the dynamics approximately follow a linear Koopman operator, enabling linear evolution in the lifted representation. Koopa~\cite{liu2023koopa} further extends this idea by constructing hierarchical Koopman operators with contextual adaptation, allowing more expressive long-term temporal modeling.

\textbf{(ii) Graph-based spatial-temporal models.}
DDGCRN~\cite{weng2023decomposition} dynamically learns graph structures from data and combines them with residual decomposition to separate normal and abnormal temporal components for improved forecasting. PDG2Seq~\cite{fan2025pdg2seq} explicitly models periodic and hidden temporal dependencies by constructing dynamic periodic graphs that evolve over time. SFGDE~\cite{ijcai2025p820} introduces a graph neural framework with learnable feedback signals inspired by control theory, which mitigates over-smoothing and enhances long-range dependency modeling.

\textbf{(iii) Sequence modeling methods.}
iTransformer~\cite{liu2023itransformer} treats multivariate time series as independent variable tokens and applies self-attention to model inter-variable dependencies. SimpleTM~\cite{chen2025simpletm} leverages wavelet-based tokenization to decompose signals into multiple frequency components and applies efficient attention for multiscale temporal modeling. TimeMixer~\cite{wang2023timemixer} decomposes time series into different temporal scales and applies feature mixing strategies to capture both short- and long-term dependencies.

\subsection{Experimental Settings}
We split all datasets into training, validation, and test sets with a ratio of 6:2:2.  All methods are trained under the same setting, using  previous five time steps to forecast the next ten steps. For the PeMS04 and PeMS08 datasets, NBDM utilizes the available control inputs. For the remaining three datasets without explicit control signals, NBDM relies on its memory-enhanced controller to infer latent control signals. We use mean absolute error (MAE) and root mean squared error (RMSE) to evaluate the performance of different models.

All experiments are conducted on Ubuntu 22.04.1 LTS with an Intel(R) Xeon(R) Gold 6330 CPU @ 2.00GHz, and  a single NVIDIA  GeForce GTX 3090 GPU. The model is optimized using Adam with an initial learning rate of 0.005. Training is performed for 200 epochs, and the learning rate is decayed every 20 epochs.

\subsection{Overall Performance}

\begin{table*}[t]
\centering
\caption{Performance comparison of baseline models and ablation study of NBDM on two given control inputs datasets. Bold denotes the best results and underline denotes the suboptimal results.}
\label{tab:results2}

\renewcommand{\arraystretch}{1.08}
\small
\setlength{\tabcolsep}{3.6pt}

\begin{tabular*}{0.85\textwidth}{@{\extracolsep{\fill}}lll|cccc|cccc}
\toprule

\multirow{2}{*}{\shortstack[c]{Type}} &
\multirow{2}{*}{\shortstack[c]{Model}} &
\multirow{2}{*}{\shortstack[c]{Metric}} &
\multicolumn{4}{c|}{PeMS04} &
\multicolumn{4}{c}{PeMS08} \\
\cmidrule(lr){4-7}
\cmidrule(lr){8-11}

& & &
1 & 4 & 7 & 10
& 1 & 4 & 7 & 10\\
\midrule

\multirow{4}{*}{Koopman}
& \multirow[c]{2}{*}{DeepKoopman} & RMSE & 36.40 & 37.31 & 38.35 & 39.67 & 33.57 & 35.02 & 37.11 & 39.94  \\
& & MAE & 22.73 & 23.64 & 24.64 & 25.88 & 21.28 & 22.56 & 24.31 & 26.60  \\
& \multirow[c]{2}{*}{Koopa} & RMSE & 31.08 & 37.72 & 45.85 & 52.89 & 23.94 & 30.77 & 38.23 & 44.41 \\
& & MAE & 19.58 & 24.55 & 30.53 & 35.97 & 15.60 & 20.18 & 25.52 & 30.30\\
\midrule

\multirow{6}{*}{Graph}
& \multirow[c]{2}{*}{DDGCRN} & RMSE & 29.05 & 31.80 & 33.36 & 35.11 & 21.88 & 25.96 & 27.11 & 28.79  \\
& & MAE & 18.33 & 20.01 & 21.12 & 22.89 & \underline{14.11} & 16.66 & 17.88 & 18.91 \\
& \multirow[c]{2}{*}{PDG2Seq} & RMSE & 29.08 & \underline{31.66} & \underline{33.21} & \underline{34.99} & \underline{21.61} & \underline{25.65} & \underline{27.06} & \underline{28.78}  \\
& & MAE & \underline{18.28} & \underline{19.98} & \underline{21.08} & \underline{22.35} & 14.15 & \underline{16.38} & \underline{17.66} & \underline{18.59} \\
& \multirow[c]{2}{*}{SFGDE} & RMSE & 32.29 & 33.80 & 38.07 & 42.05 & 22.78 & 26.43 & 30.35 & 33.51  \\
& & MAE & 19.10 & 21.42 & 24.47 & 27.43 & 14.44 & 17.15 & 19.72 & 22.06 \\
\midrule

\multirow{6}{*}{Sequence}
& \multirow[c]{2}{*}{iTransformer} & RMSE & \underline{29.04} & 35.67 & 41.86 & 47.93 & \underline{21.61} & 28.88 & 30.90 & 40.66 \\
& & MAE & 19.04 & 22.79 & 27.16 & 31.48 & \underline{14.11} & 18.67 & 22.90 & 26.75  \\
& \multirow[c]{2}{*}{SimpleTM} & RMSE & 29.20 & 35.95 & 42.64 & 48.86 & 21.74 & 29.11 & 35.52 & 40.97 \\
& & MAE & 18.38 & 22.98 & 27.68 & 32.04 & 14.17 & 18.80 & 23.15 & 27.09  \\
& \multirow[c]{2}{*}{TimeMixer} & RMSE & 29.58 & 33.04 & 35.01 & 39.49 & 22.06 & 26.44 & 29.75 & 33.10 \\
& & MAE & 18.46 & 21.00 & 23.07 & 25.44 & 14.24 & 17.29 & 19.73 & 22.14 \\
\midrule

\multirow{2}{*}{Ours}
& \multirow[c]{2}{*}{NBDM} & RMSE  & \textbf{28.89} & \textbf{31.60} & \textbf{33.09} & \textbf{34.78} & \textbf{21.57} & \textbf{25.29} & \textbf{26.91} & \textbf{28.59} \\
& & MAE & \textbf{18.24} & \textbf{19.88} & \textbf{20.97} & \textbf{22.12} & \textbf{14.08} & \textbf{16.35} & \textbf{17.23} & \textbf{18.43}  \\
\midrule

\multirow{4}{*}{Ablation}
& \multirow[c]{2}{*}{\hspace{0.3cm} -w/o BT} & RMSE & 29.21 & 33.05 & 35.44 & 37.83 & 22.14 & 26.61 & 29.13 & 31.65  \\
&  & MAE & 18.54 & 21.31 & 23.17 & 25.08 & 14.63 & 17.46 & 19.27 & 21.14  \\
& \multirow[c]{2}{*}{\hspace{0.3cm} -w/o ET} & RMSE & 29.21 & 32.37 & 34.24 & 36.13 & 21.90 & 25.94 & 28.09 & 29.92 \\
&  & MAE & 18.49 & 20.71 & 22.10 & 23.44 & 14.41 & 16.95 & 18.38 & 19.75 \\
\bottomrule
\end{tabular*}
\end{table*}

A comprehensive performance comparison  between the proposed method and the baselines is presented in Tab.~\ref{tab:results1} and Tab.~\ref{tab:results2}. The results show that NBDM achieves competitive performance across all datasets under both settings, with and without control inputs. Notably, while performance differences among models are  relatively small  at short prediction horizons, baseline models exhibit a clear and rapid accumulation of error as the forecast horizon increases. In contrast, NBDM maintains a lower level of error growth over time, highlighting its effectiveness in long-horizon forecasting.

The core advantage of NBDM lies in its explicit modeling of nonlinear dynamics through a bilinear formulation, which proves to be particularly effective. This advantage is most evident when comparing NBDM to models that rely on linear or Koopman-linearized dynamics, such as SFGDE and Koopa. Although these models perform  reasonably well at very short horizons, their errors increase more significantly as the forecasting horizon extends.  For instance, on the PeMS04 dataset at horizon 10, Koopa's RMSE (52.89) and MAE (35.97) are approximately 52\% and 62\% higher than those of NBDM (34.78 and 22.12), respectively. This pronounced error accumulation highlights the  limitation of linear approximations  in modeling nonlinear temporal dynamics over extended horizons. In contrast, the bilinear dynamics in NBDM, further enhanced by a parameterized error term, provide  a more  expressive  and stable representation, leading to  slower error growth  across horizons.

Furthermore, the results in Tab.~\ref{tab:results1} demonstrate that NBDM is specifically designed to address the practical challenge where control inputs are not available,  a scenario in which many existing models struggle. For example, on the Seoul PM$_{2.5}$ dataset at horizon 10, NBDM reduces MAE by approximately 10\% compared with the strongest baseline, iTransformer (MAE=2.12). This improvement can be attributed to two synergistic components, including a memory-enhanced controller that infers latent control signals from historical system states and control inputs, and the intrinsic integration of control inputs within the bilinear state evolution. This design enables NBDM to maintain robust forecasting performance even when external control signals are unavailable, representing a key advantage for real-world deployment.

Overall, the analysis confirms that the advantages of NBDM are consistent across datasets. When control inputs are available, NBDM achieves the best results across all metrics and horizons, demonstrating its effective use of explicit signals. In cases with missing control inputs, the model also reliably delivers top-tier performance. This robustness across different experimental settings  highlights the general applicability  of the proposed framework, which successfully unifies the modeling of explicit and implicit control effects within a coherent dynamical system.

\subsection{Case Study}

We evaluate the performance of the proposed controller through a two-stage comparative analysis. First, the proposed controller is deployed and tested on datasets containing explicit control inputs. Second, we examine its practical robustness by injecting increasing levels of noise into the data. The results validate the controller's effectiveness under both ideal and perturbed conditions.

\begin{figure}[t]
\centering
\includegraphics[width=1\columnwidth]{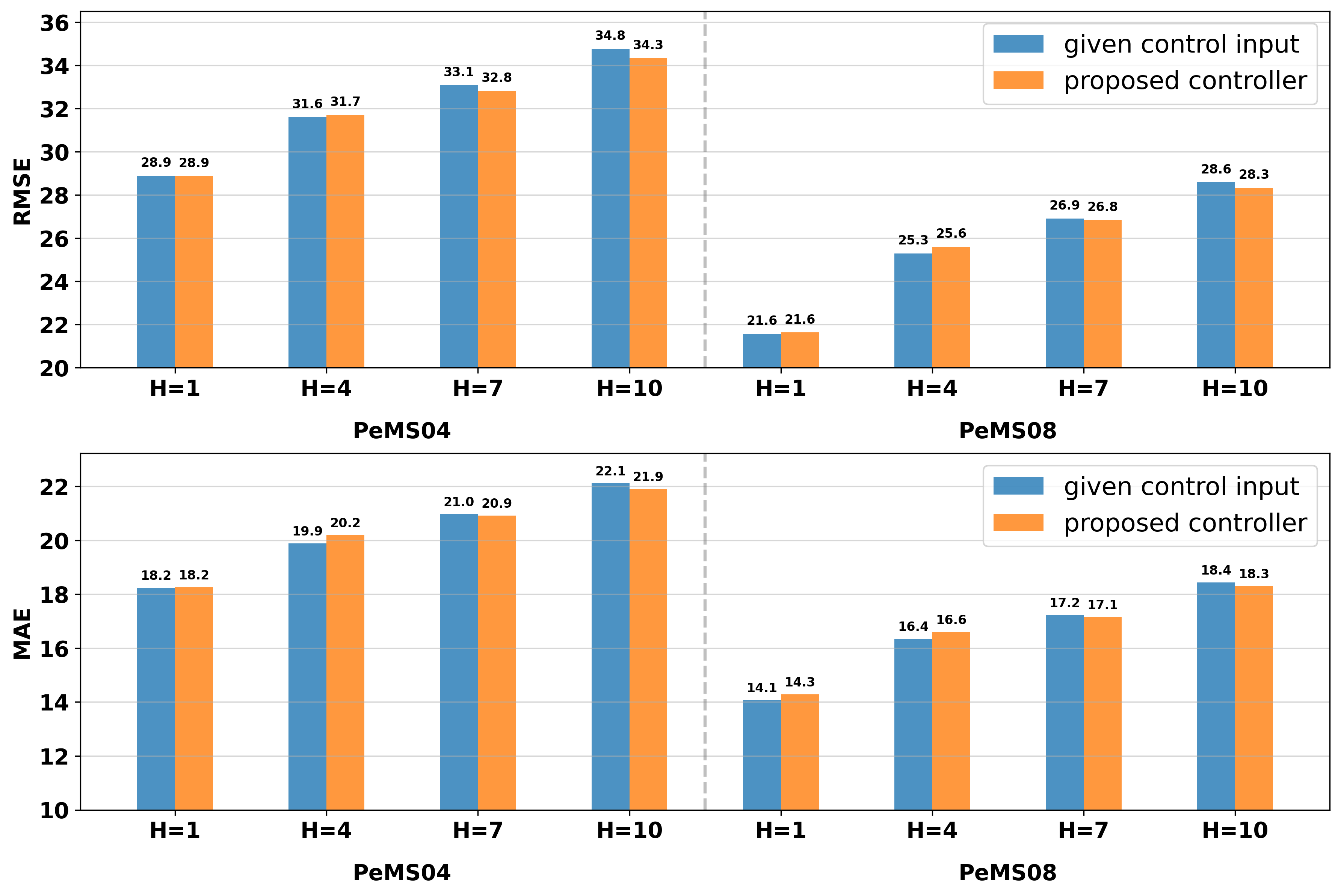}
\caption{Performance comparison of controllers using given control inputs and the proposed controllers.}
\label{fig:case_given_learn}
\end{figure}

Fig.~\ref{fig:case_given_learn} presents a performance comparison between the learned controller and the ground-truth controller. Across the provided datasets, the proposed controller achieves performance comparable to the reference controller  while consistently showing advantages in longer prediction horizons. The RMSE and MAE  of our method remain similar to those of the benchmark in all test configurations. Notably, as the prediction horizon increases, the proposed controller attains slightly better results in several scenarios, such as lower RMSE and MAE on both datasets at horizons 7 and 10. These results verify that the controller not only matches the baseline performance but also offers improved stability and accuracy in extended-range predictions, supporting its practical reliability for real-world applications.

\begin{figure}[t]
\centering
\includegraphics[width=1\columnwidth]{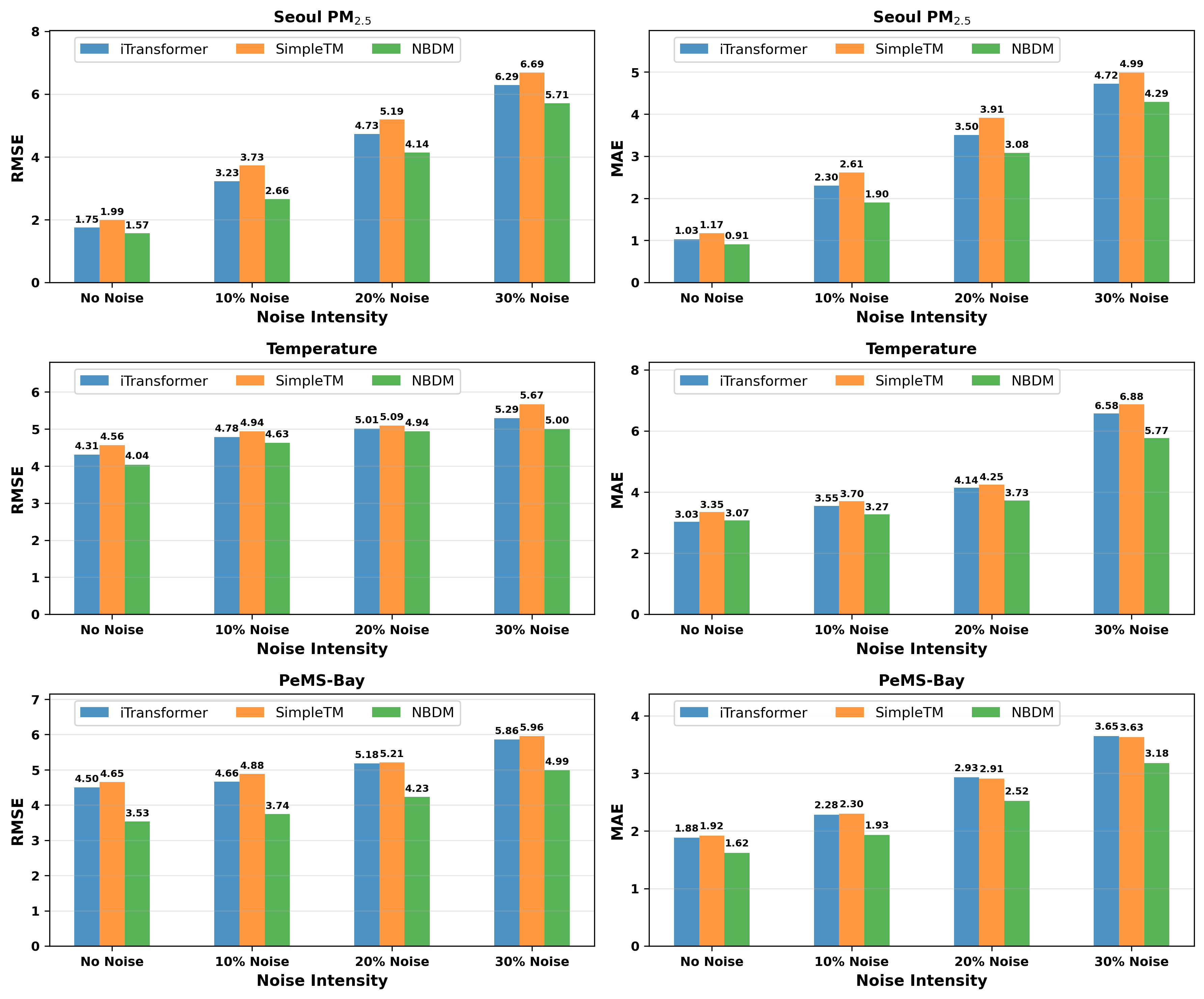}
\caption{Performance comparison of iTransformer, SimpleTM, and NBDM under varying noise levels.}
\label{fig:case_noise}
\end{figure}

The robustness of the model is further substantiated under noisy conditions. As illustrated in Fig.~\ref{fig:case_noise}, our model consistently secures the lowest RMSE and MAE across all datasets at every noise level. A key finding is that the performance advantage over baseline models, including iTransformer and SimpleTM, becomes more pronounced as noise intensity escalates. For example, on the Seoul PM$_2.5$ dataset under 30\% noise, the RMSE of our model is approximately 9.2\% and 14.6\% lower than that of iTransformer and SimpleTM, respectively. 
This widening performance gap under increasing perturbation highlights the robustness of the proposed framework. By integrating bilinear state–control interactions with feedback control, the model achieves improved stability under noisy conditions, making it well-suited for deployment in real-world environments with unreliable or corrupted observations.


\subsection{Ablation Study}

To comprehensively evaluate the contribution of each component in NBDM, we conduct an ablation study on the \textbf{B}ilinear \textbf{T}erm ($\mathbf{\tilde{B}}_t(\hat{u}_t\otimes h_t)$), the \textbf{E}rror \textbf{T}erm ($\zeta$), and the \textbf{P}roposed \textbf{C}ontroller (Eq. \ref{eq:control_input_piecewise}). Since the PeMS04 and PeMS08 datasets contain explicit control inputs, the controller ablation is performed only on datasets without available control signals. As shown in the bottom part of Tab.~\ref{tab:results1} and Tab.~\ref{tab:results2}, the full NBDM consistently achieves the best performance across datasets and forecasting horizons, indicating that each component contributes to the overall effectiveness of the model.

Removing the bilinear term leads to the largest performance degradation, particularly for long-horizon forecasting.  For example, on the Seoul PM$_2.5$ dataset at horizon 10, RMSE increases by approximately 6.0\% compared with the full model. These results suggest that explicitly modeling bilinear state-control interactions is important for capturing the nonlinear dynamics that govern system evolution over extended prediction horizons.

The removal of the error correction term also results in a consistent decrease in forecasting accuracy. The effect is particularly evident on the Temperature dataset, where RMSE increases by about 13.6\% at horizon 10. These results indicate that the error term effectively compensates for residual modeling errors arising from the bilinear formulation of nonlinear dynamics, thereby improving prediction accuracy and stability.

We further compare the proposed  controller with a conventional linear controller ($\hat{u}_t =\mathbf{W_u} h_t$).  Across all evaluated datasets and horizons, the nonlinear controller consistently achieves lower prediction errors. Moreover, the performance advantage of our controller increases as the prediction horizon extends. On the Temperature dataset, for instance, the RMSE improvement increases from 17.6\% at horizon 1 to 26.2\% at horizon 10, with the absolute error gap widening substantially. This trend indicates that the nonlinear controller not only captures more complex dynamical interactions more effectively but also exhibits improved stability and generalization in multi-step forecasting, where linear controllers often struggle to maintain accuracy.


\begin{figure}
  \centering
  \begin{subfigure}[b]{0.235\textwidth}
    \includegraphics[width=\textwidth]{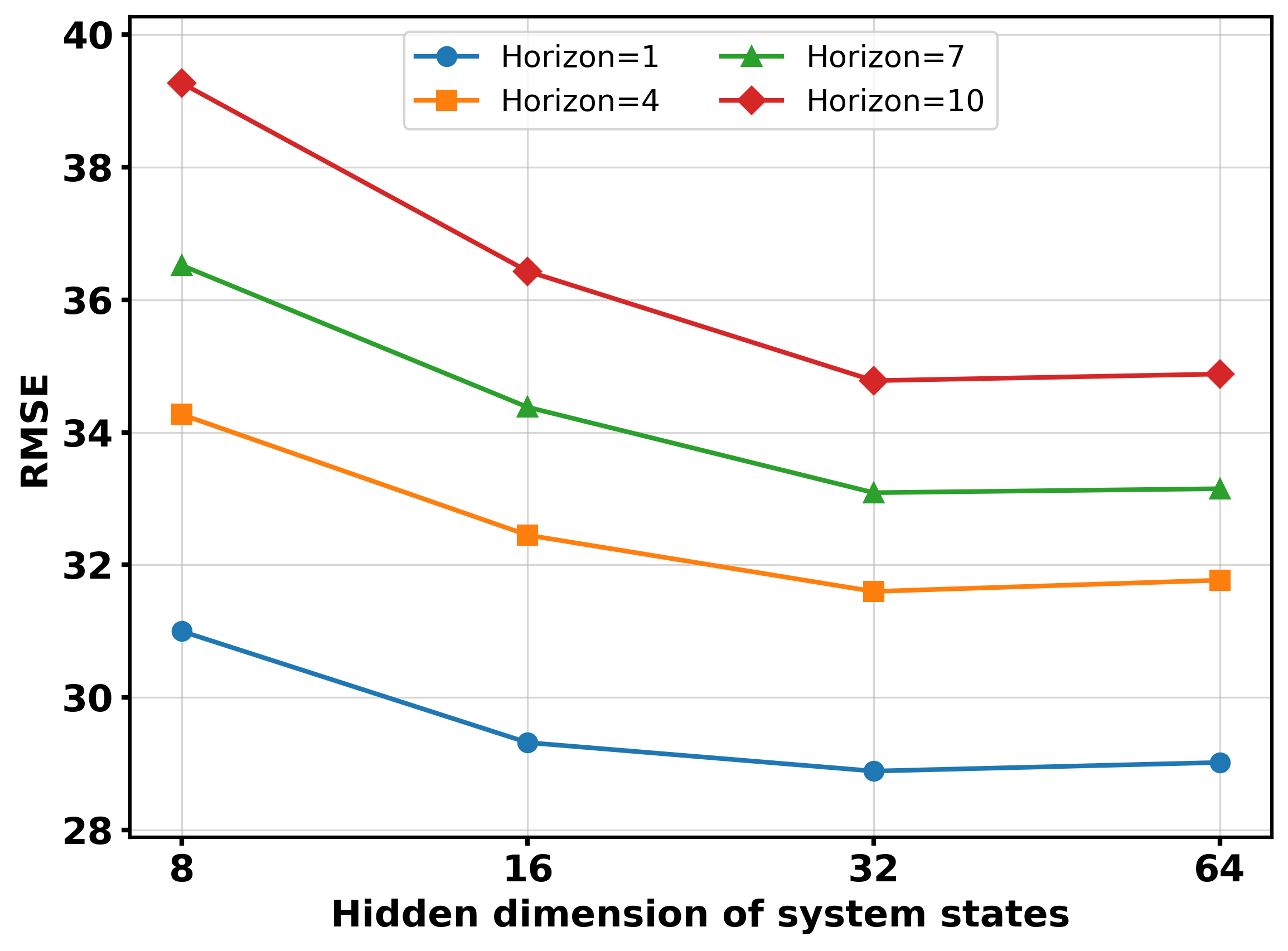}
    \label{fig:hd_rmse}
 \end{subfigure}
  \begin{subfigure}[b]{0.235\textwidth}
    \includegraphics[width=\textwidth]{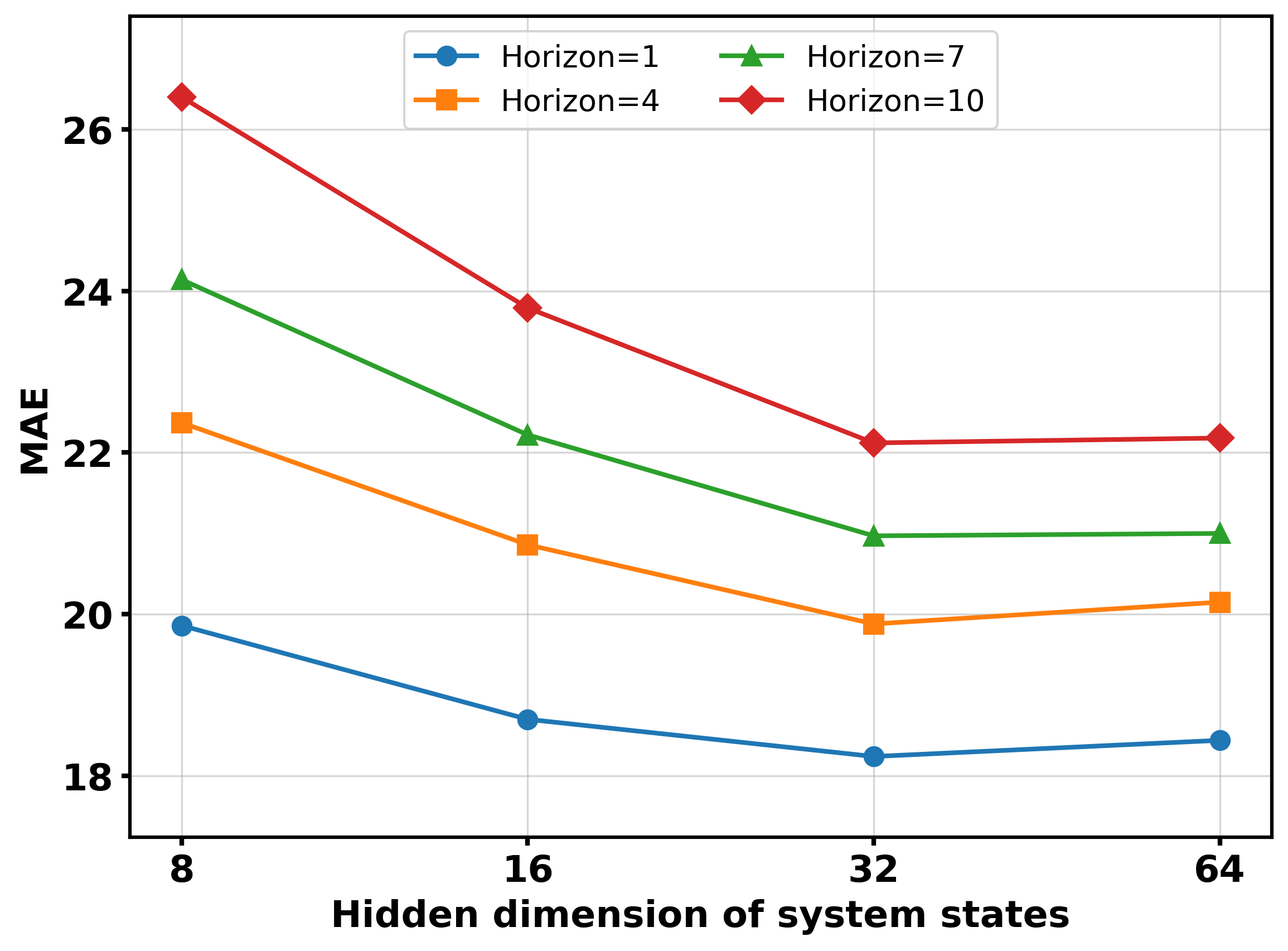}
    \label{fig:hd_mae}
  \end{subfigure}
  \begin{subfigure}[b]{0.235\textwidth}
    \includegraphics[width=\textwidth]{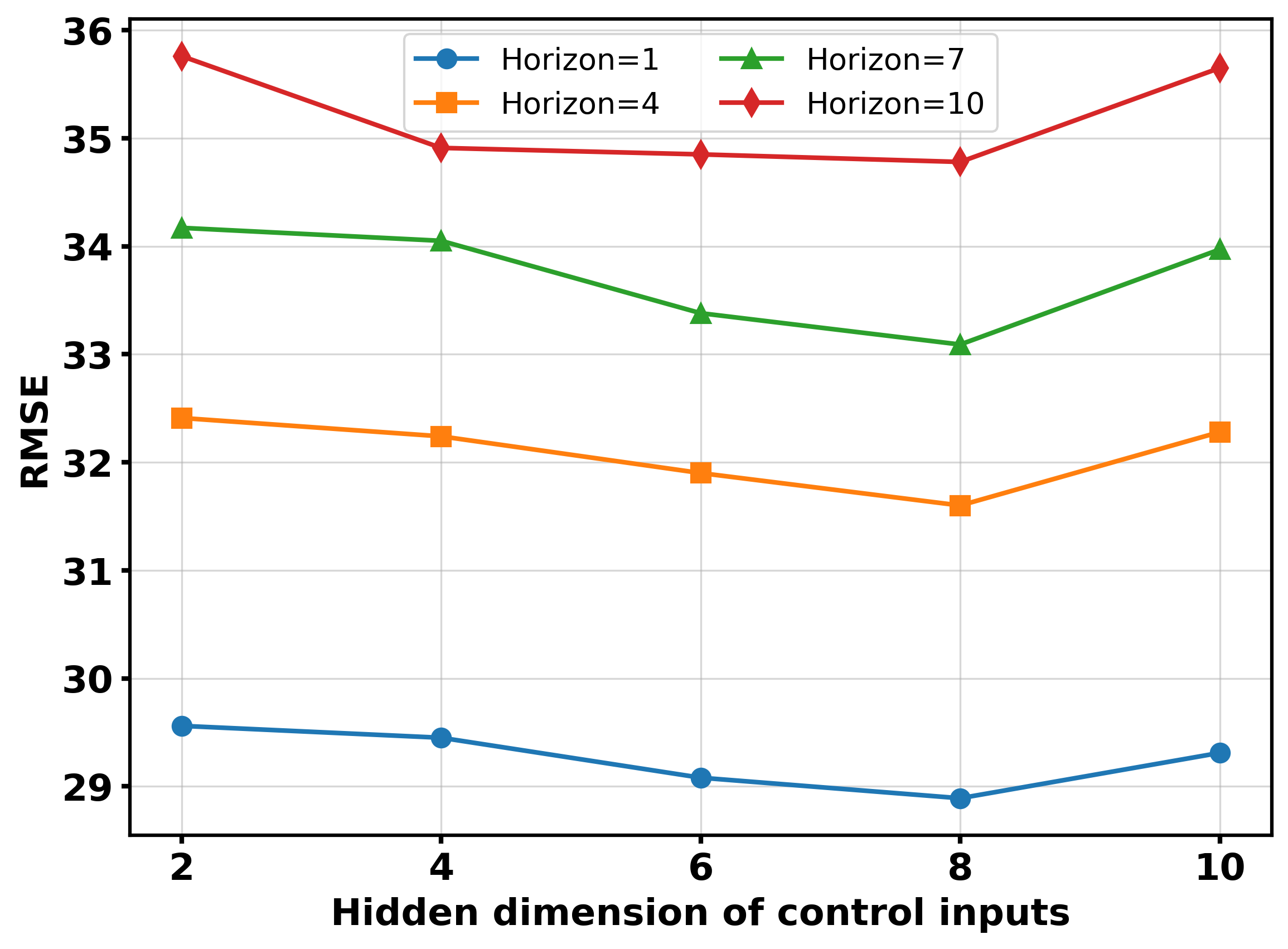}
    \label{fig:ud_rmse}
 \end{subfigure}
  \begin{subfigure}[b]{0.235\textwidth}
    \includegraphics[width=\textwidth]{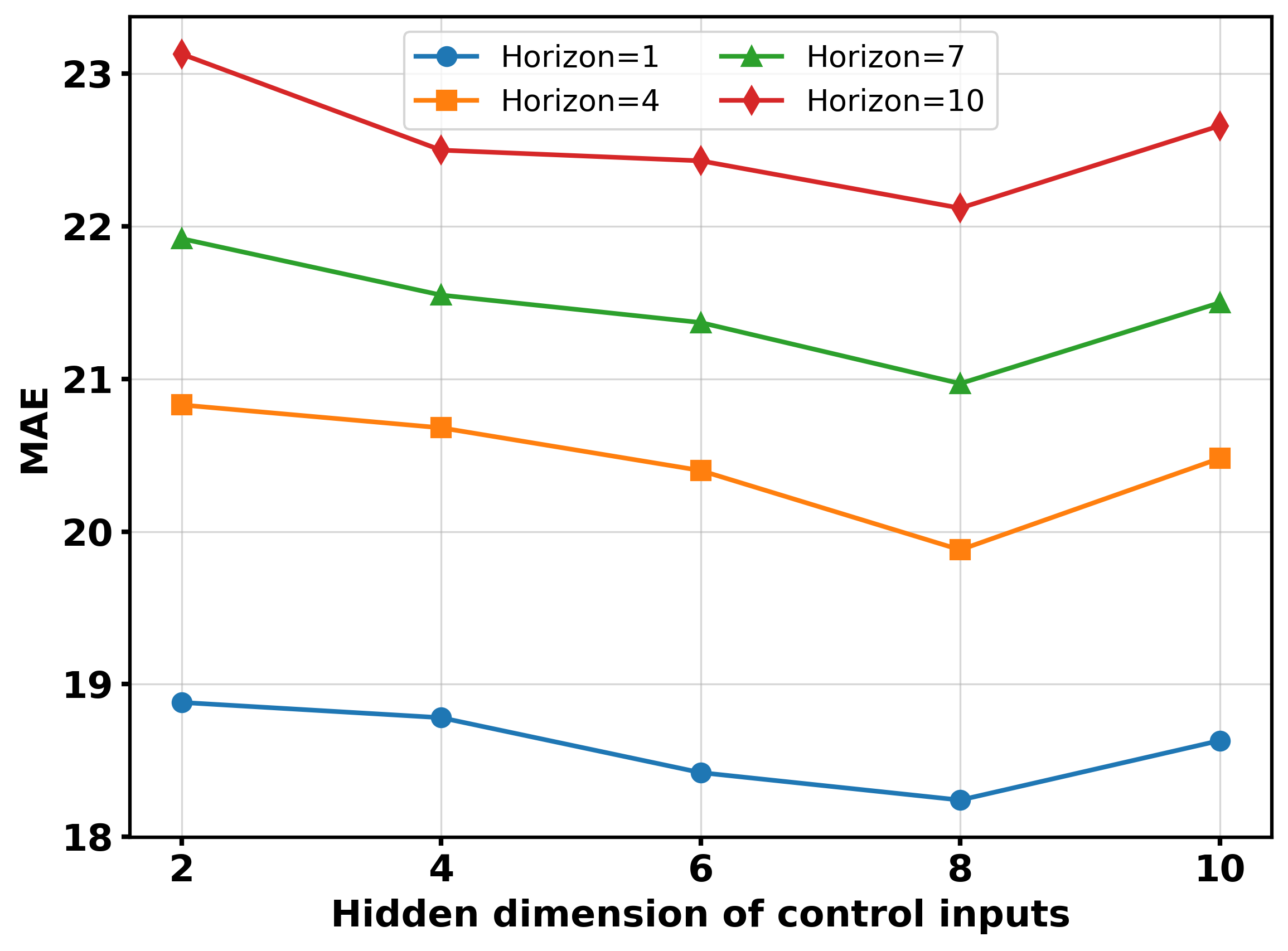}
    \label{fig:ud_mae}
  \end{subfigure}
  \caption{Parameter analysis of hidden  dimensions for system states and control inputs on PeMS04.}
  \label{fig:Parameter Analysis}
\end{figure}

\subsection{Parameter Analysis}

A comprehensive parameter analysis was performed for the hidden dimensions of system states and control inputs, with the corresponding results presented in Fig.~\ref{fig:Parameter Analysis}.
As shown, both the hidden dimension of system states and control inputs exhibit a clear influence on forecasting performance across different horizons. For 
the hidden dimension of system states, RMSE and MAE generally decrease as the dimension increases from 8 to 32, indicating improved representation capacity.  However, a further increase to 64 results in slight performance degradation, suggesting potential overfitting. Similarly, increasing the hidden dimension of control inputs from 2 to 6 yields consistent accuracy gains, while dimensions beyond 8 lead to marginal or even adverse effects, particularly at longer prediction horizons. Overall, the results suggest that moderate hidden dimensions, around 32 for system states and 6 for control inputs, strike an optimal balance between model expressiveness and generalization on the PeMS04 dataset, contributing to stable and accurate multi‑step forecasts.

\section{Conclusion}
In this paper, we study forecasting in nonlinear dynamical systems with both available and unavailable  control inputs.   Existing  methods often  rely on linear or locally linear approximations, which limit their ability to capture complex nonlinear state transitions and control effects over long horizons. To address this challenge, we propose the Neural Bilinear Dynamical Model (NBDM), a unified framework that combines Koopman-based latent representations with bilinear state-control interactions and a parameterized error correction mechanism. We further introduce a memory-enhanced controller to infer latent control inputs from historical system dynamics when control signals are unavailable. Extensive experiments on five real-world datasets demonstrate that NBDM consistently achieves superior forecasting performance under both given-control and control-unavailable settings. The results show that modeling bilinear dynamics improves multi-step prediction accuracy and robustness.

Future work will focus on extending NBDM to capture more general nonlinear dynamics beyond the current bilinear formulation, incorporating uncertainty-aware control inference mechanisms, and further improving the modeling of complex dynamical systems.


\begin{acks}
This work was supported in part by the National Natural Science Foundation of China under Grant 62372146, in part by the Zhejiang Provincial Natural Science Foundation under Grant LMS25F030011, in part by the Zhejiang Province Key R\&D Program Project under Grant No. 2025C01023, and in part by the Zhejiang Provincial Key Laboratory for Sensitive Data Security Protection and Confidentiality Management under Grant No. 2024E10048.
\end{acks}

\bibliographystyle{ACM-Reference-Format}
\bibliography{sample-base}










\end{document}